\documentclass[11pt,a4paper]{article}

\usepackage[utf8]{inputenc}
\usepackage[T1]{fontenc}
\usepackage{amsmath,amssymb,amsthm}
\usepackage{graphicx}
\usepackage{hyperref}
\usepackage{xcolor}
\usepackage{booktabs}
\usepackage{enumitem}
\usepackage{tcolorbox}
\usepackage{geometry}
\usepackage{tikz}
\usetikzlibrary{shapes.geometric, fit}
\usepackage{float}
\usepackage{caption}
\usepackage{subcaption}
\usepackage{parskip}

\definecolor{usercolor}{RGB}{204, 85, 0}  
\definecolor{aicolor}{RGB}{34, 139, 34}
\definecolor{gptcolor}{RGB}{116, 83, 170}
\definecolor{geminicolor}{RGB}{66, 133, 244}
\definecolor{problemcolor}{RGB}{0, 128, 0}  

\newtcolorbox{userbox}{
    colback=usercolor!5!white,
    colframe=usercolor!75!black,
    title=\textbf{Human},
    fonttitle=\bfseries,
    left=2mm,
    right=2mm,
    top=1mm,
    bottom=1mm
}

\newtcolorbox{problembox}{
    colback=problemcolor!5!white,
    colframe=problemcolor!75!black,
    title=\textbf{IMO 2025 Problem 6},
    fonttitle=\bfseries,
    left=2mm,
    right=2mm,
    top=1mm,
    bottom=1mm
}

\newtcolorbox{aibox}[1][AI]{
    colback=aicolor!5!white,
    colframe=aicolor!75!black,
    title=\textbf{#1},
    fonttitle=\bfseries,
    left=2mm,
    right=2mm,
    top=1mm,
    bottom=1mm
}

\newtcolorbox{gptbox}{
    colback=gptcolor!5!white,
    colframe=gptcolor!75!black,
    title=\textbf{GPT-5},
    fonttitle=\bfseries,
    left=2mm,
    right=2mm,
    top=1mm,
    bottom=1mm
}

\newtcolorbox{geminibox}{
    colback=geminicolor!5!white,
    colframe=geminicolor!75!black,
    title=\textbf{Gemini 3 Pro},
    fonttitle=\bfseries,
    left=2mm,
    right=2mm,
    top=1mm,
    bottom=1mm
}

\newtheorem{theorem}{Theorem}

\newtheorem{definition}[theorem]{Definition}

\title{\resizebox{\textwidth}{!}{\textbf{Vibe Reasoning: Eliciting Frontier AI Mathematical Capabilities}}\\[0.5em] \Large \textbf{--- A Case Study on IMO 2025 Problem 6}}

\author{
    Jiaao Wu$^{1,2,*}$ \quad Xian Zhang$^{2,\dagger}$ \quad Fan Yang$^{2}$ \quad Yinpeng Dong$^{1,\dagger}$\\[1ex]
    $^1$Tsinghua University \quad $^2$Microsoft Research\\[0.5ex]
    $^*$Work done during internship at Microsoft Research\\
    $^\dagger$Corresponding authors: \texttt{zhxian@microsoft.com}, \texttt{dongyinpeng@mail.tsinghua.edu.cn}
}

\date{December 2025}

\begin{document}

\maketitle

\begin{abstract}
\noindent
We introduce \textit{Vibe Reasoning}, a human-AI collaborative paradigm for solving complex mathematical problems. Our key insight is that frontier AI models already possess the knowledge required to solve challenging problems---they simply do not know \textit{how}, \textit{what}, or \textit{when} to apply it. Vibe Reasoning transforms AI's latent potential into manifested capability through generic meta-prompts, agentic grounding, and model orchestration. We demonstrate this paradigm through IMO 2025 Problem 6, a combinatorial optimization problem where autonomous AI systems publicly reported failures. Our solution combined GPT-5's exploratory capabilities with Gemini 3 Pro's proof strengths, leveraging agentic workflows with Python code execution and file-based memory, to derive both the correct answer (2112) and a rigorous mathematical proof. Through iterative refinement across multiple attempts, we discovered the necessity of agentic grounding and model orchestration, while human prompts evolved from problem-specific hints to generic, transferable meta-prompts. We analyze why capable AI fails autonomously, how each component addresses specific failure modes, and extract principles for effective vibe reasoning. Our findings suggest that lightweight human guidance can unlock frontier models' mathematical reasoning potential. This is ongoing work; we are developing automated frameworks and conducting broader evaluations to further validate Vibe Reasoning's generality and effectiveness.
\end{abstract}

\section{Introduction}

\subsection{Background: IMO 2025 Problem 6}

The International Mathematical Olympiad (IMO) 2025 featured an exceptionally challenging Problem 6:

\begin{problembox}
Consider a $2025 \times 2025$ grid of unit squares. Matilda wishes to place on the grid some rectangular tiles, possibly of different sizes, such that each side of every tile lies on a grid line and every unit square is covered by at most one tile. \\
Determine the minimum number of tiles Matilda needs to place so that each row and each column of the grid has exactly one unit square that is not covered by any tile.
\end{problembox}

This problem proved extremely difficult: only 6 out of approximately 600 human contestants solved it correctly~\cite{imo2025results}, and public reports indicate that \textbf{all AI systems} failed to score points on this problem~\cite{deepmind2025imo,openai2025imo,deepseek2025}.

IMO 2025 P6 is particularly valuable as an AI capability benchmark because it simultaneously challenges multiple dimensions where current AI systems struggle:

\textbf{Computational intractability meets limited data.} Natural language reasoning cannot discover the solution since the search space is too large for mental enumeration and for context windows~\cite{math2021}. Yet even with code execution, exhaustive search becomes infeasible for rather small $n$ (e.g., $n=16$), leaving only a handful of data points for pattern recognition. Furthermore, recognizing that perfect squares hold the most promising key requires selective focus that current AI systems lack.

\textbf{Construction and proof duality.} The problem demands both (1) finding the optimal answer through creative construction, and (2) proving that no better solution exists. These require fundamentally different reasoning modes---exploratory pattern-finding versus rigorous logical deduction. Not many competition problems demonstrate both difficulties simultaneously.

\textbf{Geometric intuition is essential.} The problem has rich visual structure: the ``residue block'' pattern in optimal permutations, the geometric interpretation of longest increasing/decreasing subsequences (LIS/LDS) as coordinate axes in the grid, and the spatial reasoning needed for Fooling Set construction. Current language models lack native visual/geometric intuition~\cite{alphageometry2024}.

\textbf{Deep mathematical knowledge required.} The lower bound proof requires connecting the problem to the Erd\H{o}s-Szekeres theorem, which lies in the tail of standard mathematical knowledge distributions, as well as the Fooling Set framework from communication complexity that require broad mathematical training.

\subsection{Our Contribution}

We introduce \textbf{Vibe Reasoning}, a human-AI collaborative paradigm with a clear goal: \textit{minimize human intervention to generic meta-prompts while AI performs the substantive reasoning}. Our key insight is that frontier AI models already possess all the knowledge required to solve even the most challenging problems---they simply do not know \textit{how}, \textit{what}, or \textit{when} to apply this knowledge. Vibe Reasoning can elicit AI's latent capabilities by providing lightweight human guidance and transform the reasoning protential into actual problem-solving ability.

The approach combines (1) \textbf{AI as primary reasoner}; (2) \textbf{Socratic meta-prompts}---generic directives like ``verify with code'' that prompt AI reflection without revealing solutions; (3) \textbf{agentic workflows}---Python execution for computation/verification and file-based memory for context persistence; and (4) \textbf{model orchestration}---deploying different models for exploration versus proof based on their complementary strengths.

Using this methodology, we successfully solved IMO 2025 P6, obtaining both the correct answer (\textbf{2112}) and a complete mathematical proof. Through iterative refinement across multiple attempts totaling over 24,000 lines of interaction traces, we discovered the necessity of catching errors through code verification, maintaining context through file-based memory and matching model strengths to task phases, while simultaneously, our human prompts evolved from problem-specific hints toward generic, transferable meta-prompts. We document this evolution and extract key lessons about AI capabilities, human roles, and effective collaboration principles.

\vspace{1.5em}
\noindent$\ast$ \textbf{Note:} This paper presents work in progress. We are developing an automated framework to reduce interaction frequency and human expertise requirements while preserving solution quality. We are also conducting broader evaluations on additional challenging problems to validate Vibe Reasoning's generality and effectiveness.

\section{The Vibe Reasoning Framework}

\subsection{Definition}

\begin{definition}[Vibe Reasoning]
A human-AI collaborative problem-solving paradigm where:
\begin{itemize}
    \item \textbf{AI performs substantive reasoning}: autonomously generating exploration, construction, proof, and verification.
    \item \textbf{Humans provide meta-cognitive guidance}: based on intuition rather than domain expertise, sensing when to verify, when to pivot, and when to declare success.
    \item \textbf{Generic meta-prompts}: human contributions are domain-agnostic and transferable: ``verify with code,'' ``try small cases,'' ``this approach isn't promising''.
\end{itemize}
\end{definition}

The term ``vibe'' captures the essence of AI-human collaboration and the intuitive, non-expert nature of human guidance. For example, when a human says ``this feels incomplete,'' they provide a meta-cognitive signal, not a specific mathematical fix.

\subsection{Framework Components}

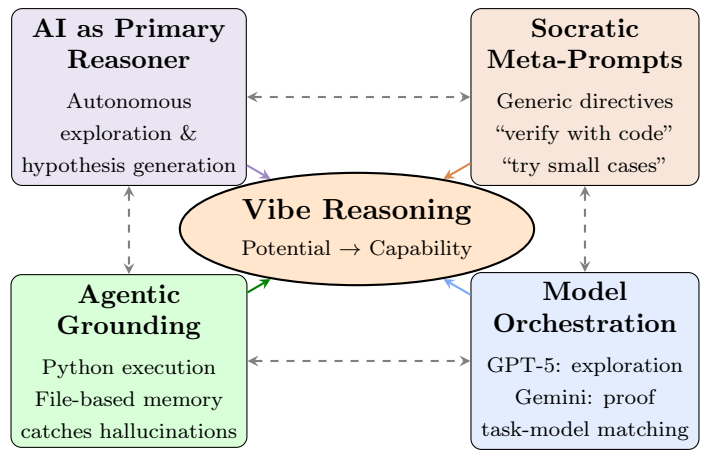
\begin{figure}[H]
\centering
\begin{tikzpicture}[
    pillar/.style={rectangle, draw, rounded corners, minimum height=2.2cm, minimum width=3cm, align=center, font=\small},
    arrow/.style={->, thick, >=stealth},
    biarrow/.style={<->, thick, >=stealth, dashed}
]
    \node[pillar, fill=gptcolor!15] (ai) at (-3, 1.5) {\textbf{AI as Primary}\\\textbf{Reasoner}\\[3pt]\scriptsize Autonomous\\\scriptsize exploration \&\\\scriptsize hypothesis generation};
    
    \node[pillar, fill=usercolor!15] (meta) at (3, 1.5) {\textbf{Socratic}\\\textbf{Meta-Prompts}\\[3pt]\scriptsize Generic directives\\\scriptsize ``verify with code''\\\scriptsize ``try small cases''};
    
    \node[pillar, fill=green!15] (agent) at (-3, -2) {\textbf{Agentic}\\\textbf{Grounding}\\[3pt]\scriptsize Python execution\\\scriptsize File-based memory\\\scriptsize catches hallucinations};
    
    \node[pillar, fill=geminicolor!15] (orch) at (3, -2) {\textbf{Model}\\\textbf{Orchestration}\\[3pt]\scriptsize GPT-5: exploration\\\scriptsize Gemini: proof\\\scriptsize task-model matching};
    
    \node[ellipse, draw, thick, fill=orange!20, minimum width=3.5cm, minimum height=1.2cm, align=center] (center) at (0, -0.25) {\textbf{Vibe Reasoning}\\\scriptsize Potential $\rightarrow$ Capability};
    
    \draw[arrow, gptcolor!70] (ai) -- (center);
    \draw[arrow, usercolor!70] (meta) -- (center);
    \draw[arrow, green!50!black] (agent) -- (center);
    \draw[arrow, geminicolor!70] (orch) -- (center);
    
    \draw[biarrow, gray] (ai) -- (meta);
    \draw[biarrow, gray] (agent) -- (orch);
    \draw[biarrow, gray] (ai) -- (agent);
    \draw[biarrow, gray] (meta) -- (orch);
\end{tikzpicture}
\caption{The Four Pillars of Vibe Reasoning: AI performs substantive reasoning, guided by generic human meta-prompts, grounded through code execution and persistent memory, with models orchestrated by complementary strengths.}
\label{fig:framework}
\end{figure}

Vibe Reasoning combines four pillars, illustrated in Figure~\ref{fig:framework}:

\textbf{(1) AI as primary reasoner}: the AI autonomously explores the problem space and generates hypotheses, rather than being guided step-by-step.

\textbf{(2) Socratic meta-prompts}: humans provide generic directives like ``verify with code'' that prompt reflection without revealing solutions. Table~\ref{tab:interventions} categorizes these transferable prompts.

\textbf{(3) Agentic grounding}: Python execution for computation/verification catches hallucinations by testing conjectures and validating constructions. File-based memory compensates for limited context windows, enabling coherent multi-session reasoning.

\textbf{(4) Model orchestration}: deploying different models for different subtasks. For instance, GPT-5 for exploration and construction, Gemini 3 Pro for rigorous proof.

\begin{table}[H]
\centering
\caption{Categories of human meta-prompts in Vibe Reasoning.}
\begin{tabular}{lp{6cm}p{4cm}}
\toprule
\textbf{Type} & \textbf{Example} & \textbf{Why Generic} \\
\midrule
Verification & ``Check with code'' & Standard practice \\
Specialization & ``Try small/special cases'' & Common heuristic \\
Challenge & ``This seems wrong'' & No specific fix given \\
Simplification & ``Summarize into one statement'' & Generic strategy \\
Output Request & ``Print the results'' & Basic debugging \\
Memory & ``Write this result to a file'' & Context management \\
\bottomrule
\end{tabular}
\label{tab:interventions}
\end{table}

\section{Case Study: Solving IMO 2025 P6}

\begin{figure}[H]
\centering
\begin{tikzpicture}[
    box/.style={rectangle, draw, rounded corners, minimum height=1cm, minimum width=2.8cm, align=center},
    smallbox/.style={rectangle, draw, rounded corners, minimum height=0.8cm, minimum width=2.2cm, align=center, font=\small},
    groupbox/.style={rectangle, draw, dashed, rounded corners, thick},
    arrow/.style={->, thick, >=stealth},
    biarrow/.style={<->, thick, >=stealth}
]
    \node[box, fill=usercolor!20] (human) at (0, 4) {Human\\\textbf{Socratic Meta-Prompts}};
    
    \node[box, fill=gptcolor!20] (gpt) at (-3, 1.5) {GPT-5: Exploration\\\textbf{AI as Primary Reasoner}};
    \node[box, fill=geminicolor!20] (gemini) at (3, 1.5) {Gemini 3 Pro: Proof\\\textbf{AI as Primary Reasoner}};
    
    \node[groupbox, black, fit=(gpt)(gemini), inner sep=8pt] (aigroup) {};
    \node[font=\small\bfseries, text=black, align=center, anchor=east] at ([xshift=-3pt]aigroup.west) {Model\\Orchestration};
    
    \node[smallbox, fill=green!20] (python) at (-2, -1.5) {Python Execution\\(Enumerate, Verify)};
    \node[smallbox, fill=yellow!30] (files) at (2, -1.5) {File-Based Memory\\(Scratch Paper)};
    
    \node[groupbox, black, fit=(python)(files), inner sep=8pt] (toolgroup) {};
    \node[font=\small\bfseries, text=black, align=center, anchor=west] at ([xshift=3pt]toolgroup.east) {Agentic Grounding};
    
    \node[box, fill=orange!20] (solution) at (0, -4) {IMO P6 Solution\\Answer: 2112 + Proof};
    
    \draw[arrow, usercolor] (human) -- node[left, font=\small, align=right] {``verify with code''\\``try small cases''} (gpt);
    \draw[arrow, usercolor] (human) -- node[right, font=\small, align=left] {``save to file''\\``check rigorously''} (gemini);
    
    \draw[biarrow, green!50!black] (gpt) -- node[left, font=\scriptsize] {execute} (python);
    \draw[biarrow, green!50!black] (gemini) -- node[right, font=\scriptsize, pos=0.5] {execute} (python);
    
    \draw[biarrow, yellow!50!black] (gpt) -- node[right, font=\scriptsize, pos=0.3] {read/write} (files);
    \draw[biarrow, yellow!50!black] (gemini) -- node[right, font=\scriptsize] {read/write} (files);
    
    \draw[arrow, gptcolor, dashed] (gpt) -- node[left, font=\small] {$M(k^2)=k^2+2k-3$} +(-2.5,-3.5) -- (solution);
    \draw[arrow, geminicolor, dashed] (gemini) -- node[right, font=\small] {Fooling Set} +(2.5,-3.5) -- (solution);
\end{tikzpicture}
\caption{Vibe Reasoning Framework instantiated for IMO P6: Human provides \textit{Socratic Meta-Prompts}; GPT-5 and Gemini 3 Pro serve as \textit{Primary Reasoners} with \textit{Model Orchestration} matching strengths to subtasks; Python execution and file-based memory provide \textit{Agentic Grounding}.}
\label{fig:imo-workflow}
\end{figure}
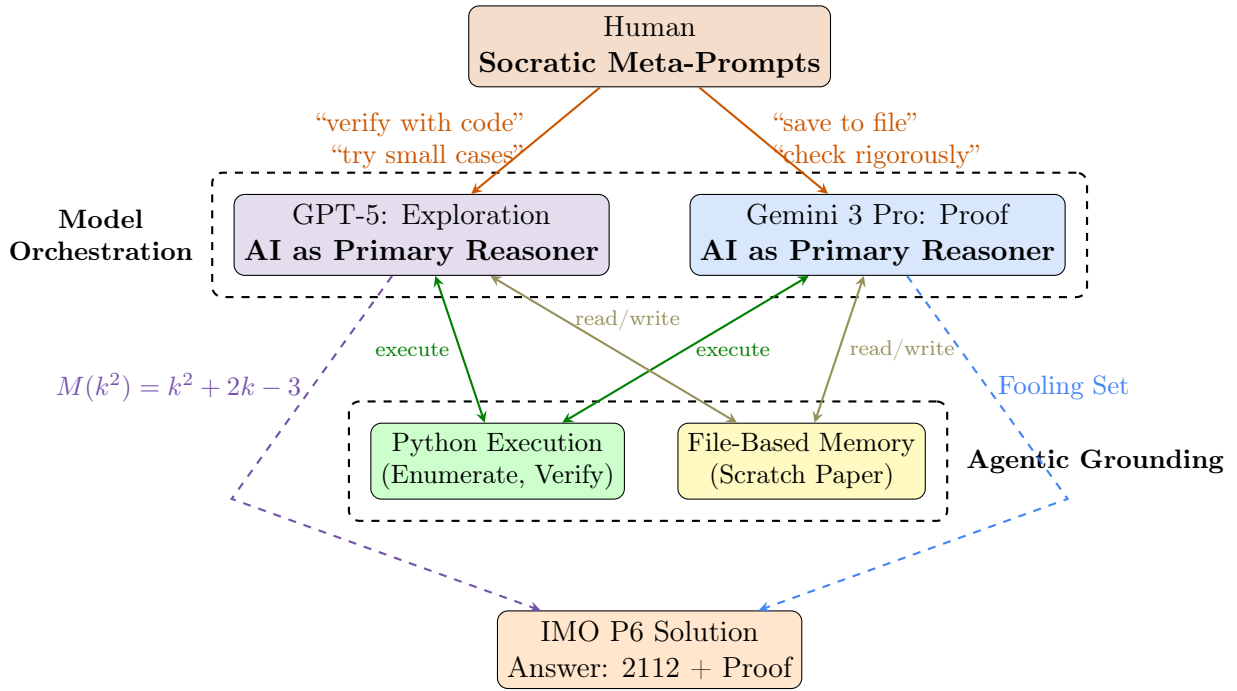

We now demonstrate Vibe Reasoning through our solution of IMO 2025 Problem 6. Figure~\ref{fig:imo-workflow} shows how all four pillars from Figure~\ref{fig:framework} instantiate for this problem. Note that different problems may require different slight variations of these components. For example, simple problems may not need model orchestration if a single model suffices.

\subsection{Phase 1: Answer Discovery (GPT-5)}
\label{sec:phase1}

The first phase focused on discovering the correct answer through exploration and enumeration. This phase primarily demonstrates: AI as primary reasoner (GPT-5 autonomously explores the problem space), Socratic meta-prompts (human guides \textit{what} to investigate and \textit{when} to verify), and agentic grounding (Python execution catches errors and enables pattern discovery).

\subsubsection{The Problem: Autonomous AI Fails}

When asked to solve the problem directly without guidance, AI produces confident but wrong answers:

\begin{figure}[H]
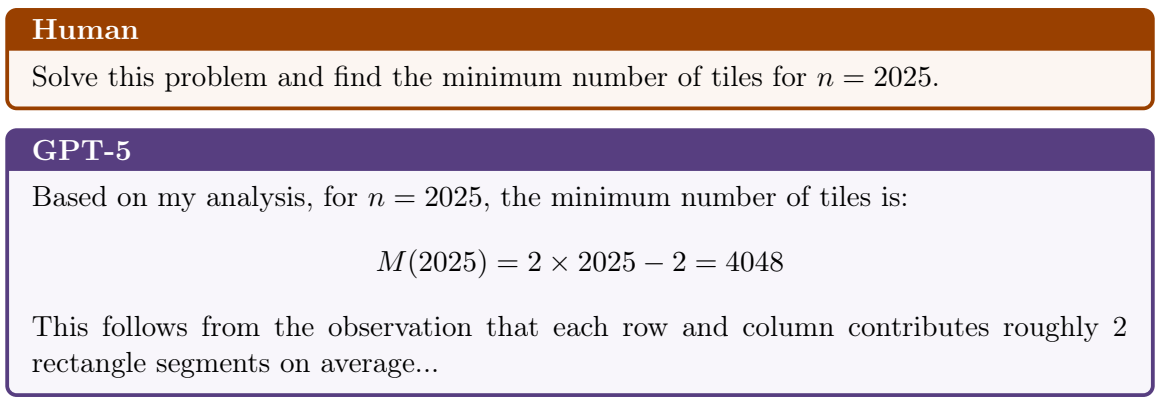

\centering
\begin{minipage}{0.95\textwidth}
\begin{userbox}
Solve this problem and find the minimum number of tiles for $n = 2025$.
\end{userbox}

\begin{gptbox}
Based on my analysis, for $n = 2025$, the minimum number of tiles is:
$$M(2025) = 2 \times 2025 - 2 = 4048$$
This follows from the observation that each row and column contributes roughly 2 rectangle segments on average...
\end{gptbox}
\end{minipage}
\caption{Autonomous attempt: confident but wrong (correct answer: 2112).}
\label{fig:overconfidence}
\end{figure}

However, with simple human guidance, the same model catches its own error:

\begin{figure}[H]
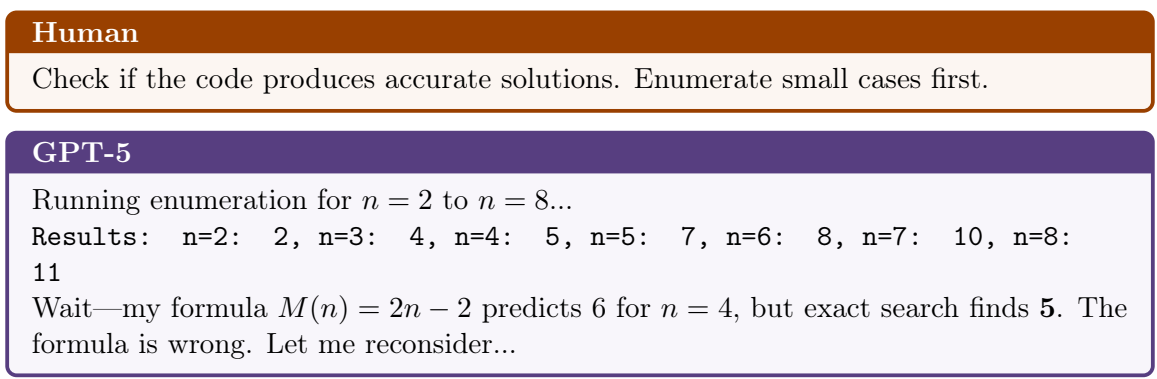

\centering
\begin{minipage}{0.95\textwidth}
\begin{userbox}
Check if the code produces accurate solutions. Enumerate small cases first.
\end{userbox}

\begin{gptbox}
Running enumeration for $n = 2$ to $n = 8$...

\texttt{Results: n=2: 2, n=3: 4, n=4: 5, n=5: 7, n=6: 8, n=7: 10, n=8: 11}

Wait---my formula $M(n) = 2n-2$ predicts 6 for $n=4$, but exact search finds \textbf{5}. The formula is wrong. Let me reconsider...
\end{gptbox}
\end{minipage}
\caption{With human guidance: same model catches its own error.}
\label{fig:verification}
\end{figure}

The human prompt in Figure~\ref{fig:verification} provides guidance on \textit{how} (``check with code'') and \textit{what} (``enumerate small cases''). Notice that the human does not suggest a specific formula or point out the error. The AI discovers the error itself once prompted to verify. This demonstrates that AI possesses the capability for self-correction; it simply lacks the judgment to know \textit{when} verification is needed. The meta-prompt triggers the AI's latent verification capability, transforming potential into actual error-catching behavior.

\vspace{1.5em}
\noindent$\ast$ \textbf{Note:} For typesetting and space considerations, the dialogue excerpts presented in the main text (above and below) have been condensed and lightly edited for clarity. They capture the essential content and flow of the interactions but may not be verbatim transcriptions. For complete, unedited conversation traces, see Appendix~\ref{appendix:proof-trace} and the supplementary materials.

\subsubsection{Initial Exploration}

\begin{figure}[H]
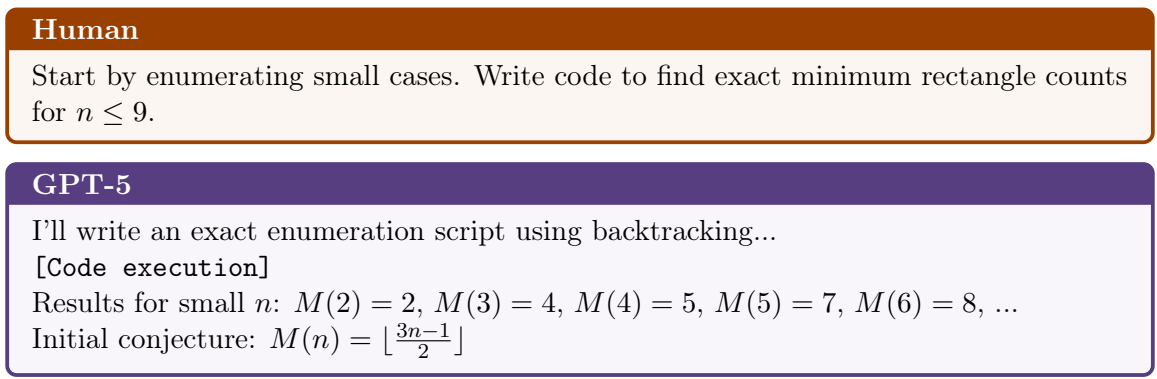

\centering
\begin{minipage}{0.95\textwidth}
\begin{userbox}
Start by enumerating small cases. Write code to find exact minimum rectangle counts for $n \leq 9$.
\end{userbox}

\begin{gptbox}
I'll write an exact enumeration script using backtracking...

\texttt{[Code execution]}

Results for small $n$: $M(2)=2$, $M(3)=4$, $M(4)=5$, $M(5)=7$, $M(6)=8$, ...

Initial conjecture: $M(n) = \lfloor \frac{3n-1}{2} \rfloor$
\end{gptbox}
\end{minipage}
\caption{Phase 1a: Small case enumeration with initial (incorrect) conjecture.}
\end{figure}

This trace demonstrates agentic grounding. The model writes and executes code to gather data, then forms a conjecture. Without code execution, it would have no reliable way to generate these data points. However, note that the AI's initial conjecture is still incorrect. It fits a linear pattern to all $n$ values. The human next provides guidance on \textit{what} specific cases to focus on.

\subsubsection{Critical Human Guidance: Focus on Perfect Squares}

\begin{figure}[H]
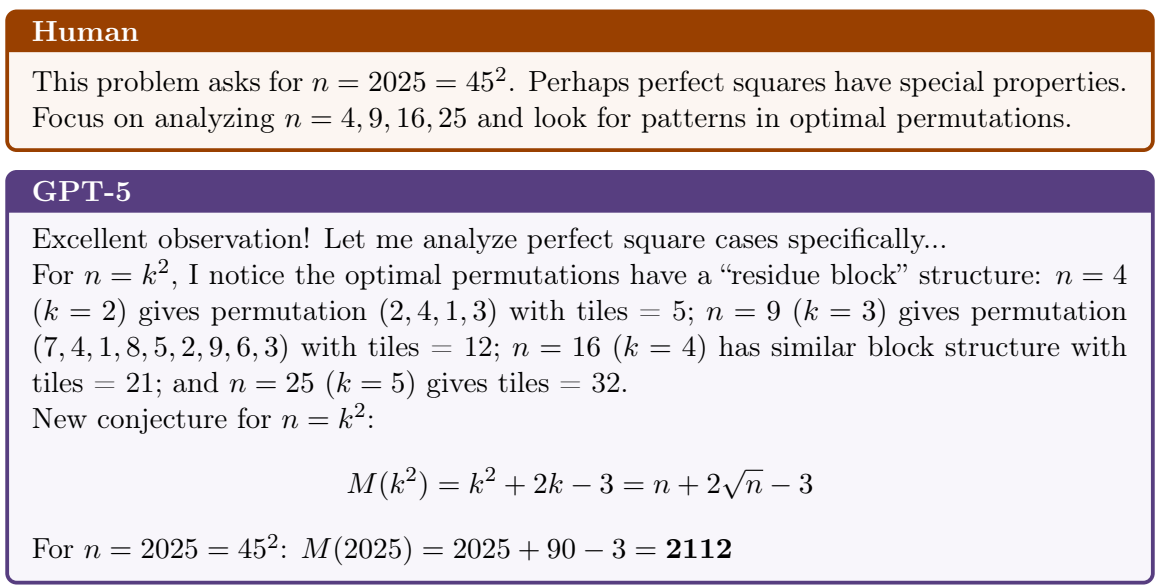

\centering
\begin{minipage}{0.95\textwidth}
\begin{userbox}
This problem asks for $n = 2025 = 45^2$. Perhaps perfect squares have special properties. Focus on analyzing $n = 4, 9, 16, 25$ and look for patterns in optimal permutations.
\end{userbox}

\begin{gptbox}
Excellent observation! Let me analyze perfect square cases specifically...

For $n = k^2$, I notice the optimal permutations have a ``residue block'' structure: $n=4$ ($k=2$) gives permutation $(2,4,1,3)$ with tiles = 5; $n=9$ ($k=3$) gives permutation $(7,4,1,8,5,2,9,6,3)$ with tiles = 12; $n=16$ ($k=4$) has similar block structure with tiles = 21; and $n=25$ ($k=5$) gives tiles = 32.

New conjecture for $n = k^2$:
$$M(k^2) = k^2 + 2k - 3 = n + 2\sqrt{n} - 3$$

For $n = 2025 = 45^2$: $M(2025) = 2025 + 90 - 3 = \mathbf{2112}$
\end{gptbox}
\end{minipage}
\caption{Phase 1b: Human guidance leads to correct pattern discovery.}
\label{fig:phase1b}
\end{figure}

Figure~\ref{fig:phase1b} shows the human guiding \textit{what} the AI should focus on. The prompt identifies that $2025 = 45^2$ is a perfect square and directs attention to this special structure. This is an example of \textit{search space pruning}---the AI has data for many $n$ values, but lacks the judgment to identify which subset holds the key insight.

Once directed, GPT-5 autonomously discovers the ``residue block'' structure and derives the correct formula. The human did not suggest this structure or the formula; AI's mathematical knowledge and pattern recognition capabilities produced these insights.

The prompt ``perhaps perfect squares have special properties'' seems to contain problem-specific information. However, this type of guidance can be framed as a general heuristic about exploring special number-theoretic cases (primes, squares, powers of 2, etc.).

\subsubsection{Verification of Construction}

\begin{figure}[H]
\centering
\begin{minipage}{0.95\textwidth}
\begin{userbox}
Show the explicit rectangle covering for these optimal permutations. Draw ASCII visualizations.
\end{userbox}

\begin{gptbox}
Here's the covering for $n=9$ with optimal permutation $(7,4,1,8,5,2,9,6,3)$:

The permutation means: hole at (row $i$, col $\pi(i)$), i.e., row 1 has hole at col 7, row 2 at col 4, row 3 at col 1, row 4 at col 8, row 5 at col 5, row 6 at col 2, row 7 at col 9, row 8 at col 6, row 9 at col 3.

\begin{verbatim}
      col: 1 2 3 4 5 6 7 8 9
         +-------------------+
  row 1  | A A A L L L . B B |  hole at col 7
  row 2  | A A A . C C C B B |  hole at col 4  
  row 3  | . D D D C C C B B |  hole at col 1
  row 4  | E D D D C C C . F |  hole at col 8
  row 5  | E D D D . G G G F |  hole at col 5
  row 6  | E . H H H G G G F |  hole at col 2
  row 7  | I I H H H G G G . |  hole at col 9
  row 8  | I I H H H . J J J |  hole at col 6
  row 9  | I I . K K K J J J |  hole at col 3
         +-------------------+
\end{verbatim}

Rectangles: A(rows 1-2, cols 1-3), B(rows 1-3, cols 8-9), etc.
Total: 12 rectangles, matching $9 + 6 - 3 = 12$.
\end{gptbox}
\end{minipage}
\caption{Phase 1c: Explicit construction verification.}
\end{figure}

The human prompts ``show explicit covering'' and ``draw ASCII visualizations.'' This guides \textit{how} the AI should verify its conjecture---through concrete, visual construction rather than abstract reasoning. The visualization serves multiple purposes: (1) it grounds the abstract formula in concrete examples, (2) it can be checked by the human for correctness, and (3) it builds confidence before moving to the proof phase.

\textbf{Phase 1 Summary.} Through this phase, we see the interplay of three pillars: the AI as primary reasoner discovered the pattern and formula; Socratic meta-prompts guided what to investigate (perfect squares) and how to verify (code, visualization); and agentic grounding through Python execution enabled both enumeration and construction validation. The human never suggested the residue block structure, the formula, or the mathematical content, only the meta-level guidance on where to look and how to check.

\subsection{Phase 2: Lower Bound Proof (Gemini 3 Pro)}

With the conjectured answer established, we switch to Gemini 3 Pro for rigorous proof. This phase demonstrates model orchestration in addition to the first three pillars. The decision to switch models reflects human judgment about \textit{when} exploration should yield to rigorous proof, and \textit{which} model is better suited for this task.

\subsubsection{Proof Strategy: Fooling Sets}

\begin{figure}[H]
\centering
\begin{minipage}{0.95\textwidth}
\begin{userbox}
Now prove that $n + 2\sqrt{n} - 3$ is a lower bound. Consider what mathematical tools could establish this.
\end{userbox}

\begin{geminibox}
I'll use the \textbf{Fooling Set} (or Cross-Free Set) method:

\textbf{Definition}: A set $S$ of cells is a \textit{Fooling Set} if for any two cells $(r_1, c_1), (r_2, c_2) \in S$, the rectangle $[r_1, r_2] \times [c_1, c_2]$ contains at least one ``hole'' (uncovered cell).

\textbf{Key Lemma}: If $S$ is a Fooling Set, then any valid tiling requires at least $|S|$ rectangles, since no rectangle can cover two cells in $S$.

\textbf{Strategy}: Construct a Fooling Set of size $n + 2\sqrt{n} - 3$.
\end{geminibox}
\end{minipage}
\caption{Phase 2a: Gemini 3 Pro proposes the Fooling Set approach.}
\end{figure}

The human prompt asks AI to ``consider what mathematical tools could establish this'' without suggesting specific techniques. Gemini 3 Pro independently proposes the Fooling Set framework, a sophisticated mathematical tool that the human did not suggest. (In the original trace, the choice of this approach was preceded by some exploration of alternative methods.) This demonstrates that AI possesses deep mathematical knowledge; the human's role is to prompt its deployment at the right moment (\textit{when}: after construction is verified) for the right purpose (\textit{what}: establishing lower bound).

\subsubsection{Constructing the Fooling Set}

\begin{figure}[H]
\centering
\begin{minipage}{0.95\textwidth}
\begin{geminibox}
The construction uses the geometric structure of LIS and LDS:

\textbf{Base Set} ($n$ cells): For each hole at $(i, \pi(i))$, select one adjacent cell.

\textbf{Augmentation} ($2\sqrt{n} - 3$ additional cells): Using the ``Orthogonal Fanning'' strategy based on the Longest Increasing Subsequence (LIS) and Longest Decreasing Subsequence (LDS). By Erd\H{o}s-Szekeres theorem, $|LIS| \cdot |LDS| \geq n$, and therefore $|LIS| + |LDS| \geq 2\sqrt{n}$. On LIS cells, we select ``outward'' in horizontal direction; on LDS cells, we select ``outward'' in vertical direction.

The ``outward'' direction ensures any two selected cells have a hole between them.
\end{geminibox}
\end{minipage}
\caption{Phase 2b: Fooling Set construction using LIS/LDS structure.}
\end{figure}

Gemini 3 Pro autonomously connects the Fooling Set construction to the Erd\H{o}s-Szekeres theorem, recognizing that LIS $\cdot$ LDS $\geq n$ implies LIS + LDS $\geq 2\sqrt{n}$, which is exactly the bound needed. This connection requires substantial mathematical knowledge and the ability to synthesize concepts from different areas (combinatorics, communication complexity). The human provides no hint about these mathematical tools; AI's role as primary reasoner generated the substantive mathematical content. But without previous interaction context, this insight might not have emerged.

\subsubsection{Verification and Completion}

\begin{figure}[H]
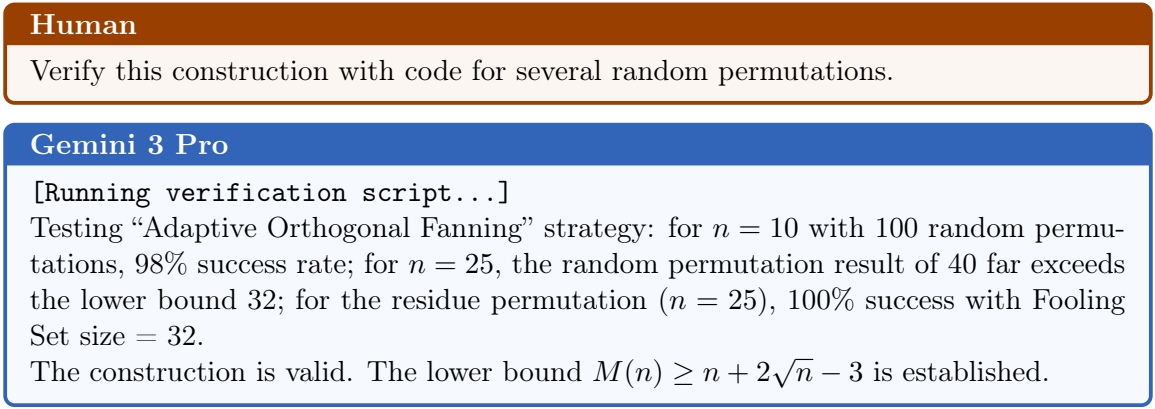

\centering
\begin{minipage}{0.95\textwidth}
\begin{userbox}
Verify this construction with code for several random permutations.
\end{userbox}

\begin{geminibox}
\texttt{[Running verification script...]}

Testing ``Adaptive Orthogonal Fanning'' strategy: for $n=10$ with 100 random permutations, 98\% success rate; for $n=25$, the random permutation result of 40 far exceeds the lower bound 32; for the residue permutation ($n=25$), 100\% success with Fooling Set size = 32.

The construction is valid. The lower bound $M(n) \geq n + 2\sqrt{n} - 3$ is established.
\end{geminibox}
\end{minipage}
\caption{Phase 2c: Computational verification of proof construction.}
\end{figure}

Even for rigorous mathematical proof, agentic grounding remains essential. The human prompts verification ``with code for several random permutations,'' and computational testing confirms the construction's validity. This illustrates a key principle: mathematical reasoning and computational verification are complementary. The AI generates the proof structure; code execution validates it.

A visualization of the Adaptive Orthogonal Fanning strategy, which we instructed Gemini 3 Pro to generate after proof, is provided in Appendix~\ref{appendix:fanning-visualization}. This aligns with the geometric construction described in Evan Chen's IMO solution~\cite{chen2025imo}, illustrating the correctness of Gemini 3 Pro's proof.

\textbf{Phase 2 Summary.} This phase demonstrates model orchestration: GPT-5 excelled at exploration but struggled with rigorous proof; Gemini 3 Pro excelled at systematic proof construction. The human's judgment about \textit{when} to switch models and \textit{which} model to use for each phase was essential. Within Phase 2, the human provided only generic guidance (``consider what tools,'' ``verify with code''), while Gemini 3 Pro independently identified the Fooling Set framework and connected it to the Erd\H{o}s-Szekeres theorem.

\subsection{Case Study Summary: Vibe Reasoning in Practice}

The IMO P6 solution demonstrates how all four pillars of Vibe Reasoning work together:

\textbf{AI as primary reasoner.} GPT-5 discovered the residue block pattern and conjectured the formula; Gemini 3 Pro identified the Fooling Set framework and constructed the proof. The human never provided mathematical content.
    
\textbf{Socratic meta-prompts.} Human guidance addressed \textit{what} (focus on perfect squares), \textit{how} (verify with code, visualize constructions), and \textit{when} (switch from exploration to proof). All prompts were generic and transferable.
    
\textbf{Agentic grounding.} Python execution enabled small-case enumeration, formula validation, construction visualization, and proof verification. Without code, the AI would have accepted incorrect conjectures.
    
\textbf{Model orchestration.} GPT-5's exploratory strength found the answer; Gemini 3 Pro's rigorous strength proved the bound. Neither model alone could have achieved the complete solution.

The contrast between autonomous failure (Figure~\ref{fig:overconfidence}) and collaborative success illustrates that Vibe Reasoning can provide AI's missing guidance through lightweight, generic meta-prompts.

\section{Evolution and Lessons Learned}

Our journey to solving IMO P6 demonstrates that Vibe Reasoning is an evolving practice that improves through iteration. Table~\ref{tab:timeline} summarizes our progression across multiple separate sessions, totaling over 24,000 lines of interaction traces.

\begin{table}[H]
\centering
\caption{Timeline of our attempts, showing progressive refinement of methodology.}
\begin{tabular}{lp{5.5cm}p{5cm}}
\toprule
\textbf{Phase} & \textbf{Attempt} & \textbf{Progress} \\
\midrule
1 & Initial trials with multiple models & Lean formalization failed; ineffective prompts and models led to inconsistent results. \\
2 & GPT-5 answer discovery & Success after self-refined code and discovered key structure. \\
3 & Multiple proof attempts & GPT-5 proof failures; identified model specialization need. \\
4 & GPT-5 whole trace reproduction & Answer reproduced; proof terminology overload. \\
5 & Gemini 3 Pro proof success & Complete solution achieved. \\
\bottomrule
\end{tabular}
\label{tab:timeline}
\end{table}

Each methodological improvement below emerged from specific failures and is illustrated with concrete evidence from our traces.

\subsection{Discovering the Need for Agentic Grounding}

Section~\ref{sec:phase1} demonstrated how Python execution enables pattern discovery and catches hallucinations. Here we focus on a complementary discovery: the need for file-based memory.

\textbf{Early approach}: All context remained in the conversation, leading to lost information and repeated mistakes.

\textbf{Improved approach}: AI maintains structured files (\texttt{summary.md}, \texttt{proof\_sketch.md}) as external memory.

\textbf{Evidence}: File reference counts across traces show adoption of this practice:
\begin{itemize}
    \item \textbf{Early exploratory trace}: Minimal file usage, frequent context loss across model switches.
    \item \textbf{Successful answer trace}: $\sim$106 file references, systematic use of \texttt{summary.md} for recording confirmed results.
    \item \textbf{Successful proof trace}: $\sim$43 file references, more focused but still essential for tracking proof attempts and avoiding repetition.
\end{itemize}

Python execution catches errors in real-time; file-based memory maintains coherence across sessions. Together, they form Agentic Grounding, external mechanisms that compensate for AI's unreliable self-evaluation and limited context windows.

\subsection{Discovering the Need for Model Orchestration}

\textbf{Early approach}: We attempted to use a single model for the entire problem.

\textbf{Improved approach}: After observing complementary failure modes, we recognized that GPT-5 excels at exploration while Gemini 3 Pro excels at rigorous proof, and deliberately switched models between phases.

\textbf{Evidence}: Comparing the GPT-5-only trace versus the multi-model success trace demonstrates this clearly:
\begin{itemize}
    \item \textbf{GPT-5-only trace}: Answer discovery succeeded, but proof attempts failed repeatedly with schemes like ``Guard Scheme'' and ``Quadrant Coloring'' producing 100\% violation rates.
    \item \textbf{Gemini 3 Pro proof trace} (given answer from GPT-5): Achieved complete proof with Fooling Set approach.
\end{itemize}

\begin{figure}[H]
\centering
\begin{minipage}{0.95\textwidth}
\begin{gptbox}
Let me try the ``Guard Scheme'' approach...

\texttt{[Code verification]}

\textbf{Result}: Violations found in 100\% of test cases.

The guard scheme fails because vertical cells placed at $g-1$ instead of $g$ break the piercing property...
\end{gptbox}
\end{minipage}
\caption{GPT-5's proof attempts failed despite extensive effort, motivating the switch to Gemini 3 Pro.}
\end{figure}

Different models have complementary strengths; no single model excels at all phases. This motivated Model Orchestration. The human's role in recognizing when to switch models and which model suits each phase requires no problem-specific knowledge, only meta-level judgment.

\subsection{The Evolution Toward Generic Meta-Prompts}

Across all our attempts, we observed a clear trend: earlier traces required more problem-specific human guidance, while later successful attempts relied increasingly on generic meta-prompts. This evolution was not intentional at first, but emerged as we discovered what actually worked. We illustrate this evolution across three categories of prompts.

\subsubsection{Verification Prompts}

\textbf{Early approach}: AI proposed conjectures and we accepted seemingly plausible justifications without systematic verification.

\textbf{Improved approach}: We developed generic verification prompts that catch errors without requiring domain knowledge.

\textbf{Evidence}: Comparing two GPT-5 answer discovery attempts reveals that success depends on \textit{when} verification occurs:
\begin{itemize}
    \item \textbf{Successful trace}: The human prompted ``check if the code produces accurate solutions'' \textit{early}, before pattern-fitting. This caught a subtle enumeration bug that would have corrupted all subsequent observations.
    \item \textbf{Failed trace}: The human accepted initial code output without verification prompts. A bug went undetected, leading to wrong conclusions.
\end{itemize}
The key difference was a single well-timed meta-prompt, entirely generic and transferable to any problem involving code.

\subsubsection{Role-Switching Prompts}

\textbf{Early approach}: Seemingly plausible AI reasoning was accepted without challenge.

\textbf{Improved approach}: Explicit role-switching prompts (``you are an extremely strict reviewer'') to catch errors.

\textbf{Evidence}: The Gemini answer failure traces illustrate what happens without such prompts:
\begin{figure}[H]
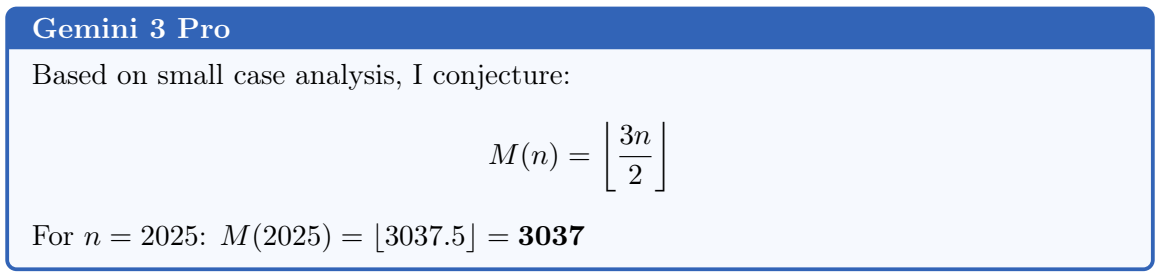

\centering
\begin{minipage}{0.95\textwidth}
\begin{geminibox}
Based on small case analysis, I conjecture:
$$M(n) = \left\lfloor \frac{3n}{2} \right\rfloor$$

For $n = 2025$: $M(2025) = \lfloor 3037.5 \rfloor = \mathbf{3037}$
\end{geminibox}
\end{minipage}
\caption{Gemini 3 Pro's incorrect conjecture. It accepted linear pattern without verifying against perfect square structure.}
\end{figure}
In contrast, the successful trace used prompts like ``focus only on perfect square cases'' that forced verification.

\subsubsection{Planning and Exploration Prompts}

\textbf{Early attempts}: Human prompts contained problem-specific hints:
\begin{itemize}
    \item ``You mentioned the Erd\H{o}s--Szekeres theorem, how might we use it?'' (suggesting a specific mathematical tool)
    \item ``The holes form a permutation matrix, try thinking about LIS/LDS'' (revealing structural insight)
    \item ``Consider cells adjacent to holes as candidates for the Fooling Set'' (guiding proof construction)
\end{itemize}

\textbf{Later successful attempts}: The same results were achieved with generic prompts:
\begin{itemize}
    \item ``First plan your approach, propose 5 different strategies'' (generic planning)
    \item ``Write code to verify your construction'' (generic verification)
    \item ``Focus on special cases---what makes $n=2025$ special?'' (generic heuristic about special structure)
\end{itemize}

The successful proof trace achieved the complete solution without the human suggesting the Fooling Set framework, the Erd\H{o}s-Szekeres theorem, or the LIS/LDS connection. The AI independently identified these from a menu of self-generated options. The human's role was purely meta-cognitive: directing \textit{when} to verify, \textit{what} categories to explore, and \textit{how} to structure the search.

This evolution demonstrates a key observation of Vibe Reasoning: Generic human guidance can be effective. The human provides meta-cognitive direction (\textit{how}, \textit{what}, \textit{when}) while AI supplies the domain knowledge. As we refined our methodology, we converged on Socratic Meta-Prompts: transferable prompts that work across problems. This also enlightens our future work in automated frameworks.

\subsection{Summary: Why Autonomous AI Fails}
\label{sec:lessons}

The methodological improvements above were hard-won through failure. Each lesson corresponds to a specific AI weakness that the problem exposed:

\textbf{The Knowledge-Application Gap.} AI models possess extensive mathematical knowledge---they can state the Erd\H{o}s-Szekeres theorem and describe Fooling Set constructions. Yet when solving IMO P6 directly, they fail. The problem is not missing knowledge but missing judgment about when and how to apply it.

\textbf{Overconfidence and Verification Blindness.} Models present incorrect answers (e.g., $M(n) = 2n-2$) with high confidence, lacking self-awareness to recognize when claims need verification. They rarely spontaneously verify claims with code. Even when verification is possible, they default to reasoning-based confirmation sharing the same blind spots as the original claim. Once a formula is proposed, they defend rather than question it.

\textbf{Circular Proof Attempts.} In proof phases, AI cycled through variations of the same failed approaches (guard cells $\rightarrow$ quadrant coloring $\rightarrow$ parity augmentation) without recognizing the need for fundamentally different strategies. They lack meta-cognitive insight to evaluate the overall proof strategy and pivot when necessary.

\textbf{Context Loss and Exploration-Rigor Trade-off.} Without persistent memory, insights from failed attempts are lost across sessions. Models excelling at creative exploration may struggle with formal proofs, while rigorous provers sometimes miss exploratory insights.

\textbf{The Core Issue: Unreliable Self-Evaluation.} All these failure modes share a common root: AI cannot reliably evaluate its own outputs. The same model expresses identical confidence for correct and incorrect answers. Without external grounding such as code execution or human judgment, AI cannot distinguish insight from hallucination. This explains why human guidance, even minimal and non-technical, proves essential: it provides the independent evaluation that AI lacks.

\section{Discussion}
\label{sec:discussion}

Section~\ref{sec:lessons} identified why autonomous AI fails on complex problems like IMO P6. We now discuss how each pillar of Vibe Reasoning addresses these challenges, following the framework structure from Section 2.2.

\subsection{Pillar 1: AI as Primary Reasoner}

Vibe Reasoning fundamentally depends on sufficiently capable foundation models. The AI must be the \textit{primary reasoner}---autonomously exploring, hypothesizing, and constructing---rather than being guided step-by-step. Several capabilities proved essential:

\textbf{Pattern Recognition}: The ability to observe patterns in small-$n$ enumeration results requires strong inductive reasoning.
    
\textbf{Mathematical Knowledge Association}: Connecting the problem to relevant theorems, such as recognizing that the Erd\H{o}s-Szekeres theorem provides exactly the bound needed, requires broad mathematical training.
    
\textbf{Code Competence}: Writing correct enumeration and verification code is non-trivial; early traces show that buggy code led to incorrect conclusions.

Weaker or insufficiently trained models cannot support Vibe Reasoning. They lack the baseline capabilities to generate meaningful hypotheses or execute correct verification. This represents a \textit{prerequisite} rather than a contribution of our methodology.

\subsection{Pillar 2: Socratic Meta-Prompts and Human Guidance}

As Section~\ref{sec:lessons} established, AI's core limitation is unreliable self-evaluation. The human's role is to provide the independent judgment that AI lacks---not domain expertise, but \textit{meta-cognitive guidance} on \textit{how}, \textit{what}, and \textit{when} to apply AI's existing knowledge.

\subsubsection{How: Out-of-Distribution Guidance}

When AI encounters situations outside its training distribution, it defaults to familiar but inappropriate patterns. Humans recognize these moments and redirect:

\begin{itemize}
    \item \textbf{Recognizing special structure}: When AI fitted linear patterns to all $n$, the human prompted ``focus on perfect squares since $2025 = 45^2$.'' This out-of-distribution insight was invisible to the AI but obvious to the human examining the data.
    
    \item \textbf{Choosing representations}: The human recognized that LIS and LDS could serve as natural ``coordinate axes'' for the Fooling Set construction. The AI persistently tried row/column indices, requiring correction.
    
    \item \textbf{Breaking fixation}: When AI circled through failed proof approaches (``guard cells,'' ``quadrant coloring,'' ``parity augmentation''), the human prompted a fundamentally different direction rather than variations on the same theme.
\end{itemize}

\subsubsection{What: Search Space Pruning and Phase Transitions}

AI can explore vast search spaces but struggles to prioritize. The human provides crucial filtering:

\begin{itemize}
    \item \textbf{Selecting promising directions}: From AI's enumeration of small cases, the human identified which patterns merited deeper investigation (residue blocks) versus dead ends (symmetric constructions).
    
    \item \textbf{Deciding phase transitions}: The judgment that ``we have the answer, now focus on proof'' requires recognizing when exploration should yield to verification. AI lacks reliable confidence calibration for such decisions.
    
    \item \textbf{Model selection}: Recognizing that GPT-5 excels at exploration while Gemini 3 Pro excels at rigorous proof, and switching at the right moment are meta-level decisions beyond current AI capability.
\end{itemize}

\subsubsection{When: Timing of Interventions}

Knowing \textit{when} to intervene is perhaps the subtlest human contribution:

\begin{itemize}
    \item \textbf{Letting productive exploration continue}: Not every AI tangent should be interrupted. The human must judge when to let AI pursue a direction versus when to redirect.
    
    \item \textbf{Recognizing diminishing returns}: After three failed proof attempts using similar techniques, the human recognized the need for a fundamentally different approach.
    
    \item \textbf{Cross-session coherence}: Unlike humans who maintain context across days, AI starts fresh each session. The human ensures insights from failed attempts inform later ones through structured notes and strategic prompting.
\end{itemize}

Can these human judgments be automated? The core obstacle, as analyzed in Section~\ref{sec:lessons}, is that AI lacks reliable self-evaluation. Beyond this, meta-cognitive oversight requires independence: using AI to verify AI suffers from correlated errors, as both may share the same blind spots. The human provides genuinely independent assessment that current AI architectures cannot replicate. However, some aspects may be automatable in future work through \textit{roll-out strategies} (autonomous exploration of multiple special cases) or \textit{ensemble verification} (using diverse models to cross-check claims).

\subsection{Pillar 3: Agentic Grounding through Python Execution}

Python code execution was not merely helpful but \textit{essential} for success. As Section~\ref{sec:lessons} showed, AI reasoning without grounding remains prone to hallucination.

\subsubsection{Catching Hallucinations Early}

The most important role of code execution was catching confident but incorrect claims:

\begin{itemize}
    \item \textbf{Formula Verification}: GPT-5 initially claimed $M(n) = 2n - 2$. Running enumeration code immediately showed $M(4) = 5 \neq 6$, forcing reconsideration.
    
    \item \textbf{Construction Validation}: When GPT-5 proposed rectangle coverings, code verified whether they actually satisfied the constraints (each row/column has exactly one hole).
    
    \item \textbf{Proof Checking}: Gemini 3 Pro's Fooling Set construction was tested on random permutations. Early versions had bugs; code caught violations that pure reasoning missed.
\end{itemize}

\subsubsection{Enabling Pattern Discovery}

Beyond verification, code execution enabled discoveries that pure reasoning could not achieve:

\begin{itemize}
    \item \textbf{Exhaustive Search}: For $n \leq 8$, code enumerated all valid configurations to find exact minima---something impractical by hand.
    
    \item \textbf{Optimal Permutation Analysis}: Code identified that the permutation $(7,4,1,8,5,2,9,6,3)$ for $n=9$ achieves the minimum. Printing this permutation revealed the residue block structure.
    
    \item \textbf{Formula Fitting}: Given data points $(4,5), (9,12), (16,21), (25,32)$, code fit the pattern $M(k^2) = k^2 + 2k - 3$.
\end{itemize}

\subsection{Pillar 4: Model Orchestration}

Our experiments revealed that different models have distinct strengths, and \textbf{model orchestration}---using different models for different phases---is crucial for complex problems.

\begin{table}[H]
\centering
\begin{tabular}{lcc}
\toprule
\textbf{Capability} & \textbf{GPT-5} & \textbf{Gemini 3 Pro} \\
\midrule
Pattern Exploration & \textcolor{green!60!black}{\textbf{Strong}} & Moderate \\
Code-Based Verification & \textcolor{green!60!black}{\textbf{Strong}} & Moderate \\
Construction Discovery & \textcolor{green!60!black}{\textbf{Strong}} & Weak \\
Rigorous Proof & Weak & \textcolor{green!60!black}{\textbf{Strong}} \\
Systematic Verification & Moderate & \textcolor{green!60!black}{\textbf{Strong}} \\
\bottomrule
\end{tabular}
\caption{Model capabilities comparison on IMO P6 tasks.}
\label{tab:model-comparison}
\end{table}

GPT-5 demonstrated exceptional ability in open-ended exploration and pattern discovery. When given the meta-prompt ``enumerate small cases,'' it autonomously wrote backtracking code to find exact minimum tile counts for $n \leq 9$, identified that optimal configurations correspond to permutations, discovered the residue block structure (grouping positions by $i \mod k$ and arranging blocks in descending order), and conjectured the formula $M(k^2) = k^2 + 2k - 3$ by fitting patterns. However, GPT-5 struggled with rigorous proofs. Its attempts at proving the lower bound produced increasingly baroque schemes (``guard cells,'' ``quadrant coloring,'' ``parity augmentation'') that failed verification.

Gemini 3 Pro showed the opposite profile. When asked to find the answer directly, it converged on the incorrect formula $M(n) = \lfloor 3n/2 \rfloor$, giving 3037 instead of 2112. It failed to discover the special structure of perfect squares despite extensive prompting. However, once given the correct answer and asked to prove it, Gemini 3 Pro excelled. It immediately identified the Fooling Set / Cross-Free Set framework as appropriate, connected the problem to the Erd\H{o}s-Szekeres theorem on LIS and LDS, developed the ``Orthogonal Fanning'' construction systematically, and verified the construction with careful case analysis.

In summary, Vibe Reasoning resolves the paradox of capable-yet-failing AI, namely frontier models have the knowledge but lack the judgment to deploy it effectively. The four pillars work together synergistically:
\textbf{AI as primary reasoner} provides the substantive mathematical work; \textbf{Socratic meta-prompts} supply the missing judgment on how, what, and when; \textbf{Agentic grounding} catches hallucinations and enables discovery; \textbf{Model orchestration} matches model strengths to task requirements. This transforms AI's latent potential into manifested capability, achieving results neither humans nor AI could reach alone.

\section{Related Work}

Our work builds upon several lines of research in AI-assisted mathematical reasoning and human-AI collaboration.

\textbf{AI for Mathematics.} Recent advances in large language models have enabled impressive mathematical reasoning capabilities. Systems like AlphaProof~\cite{alphaproof2024} and AlphaGeometry~\cite{alphageometry2024} have achieved medal-level performance on mathematical olympiad problems. At IMO 2025, multiple AI systems attempted the competition: DeepMind reported gold-medal-level performance overall~\cite{deepmind2025imo}, while OpenAI~\cite{openai2025imo} and DeepSeek~\cite{deepseek2025} also released their attempts. However, all systems failed on Problem 6, which required both creative exploration and rigorous proof construction. Our work demonstrates that human-AI collaboration can succeed where fully autonomous systems fail.

\textbf{Tool-Augmented Language Models.} The integration of external tools with language models has proven essential for reliable reasoning~\cite{toolformer2023,chameleon2023}. Python execution, in particular, provides crucial grounding for mathematical claims~\cite{pal2023,mathcoder2023}. Our agentic workflow extends this paradigm by combining code execution with persistent file-based memory, enabling AI to maintain context across complex, multi-phase reasoning.

\textbf{Human-AI Collaboration.} Prior work has explored various collaboration paradigms, from AI assistants that help humans~\cite{copilot2021} to humans verifying AI outputs~\cite{humanintheloop2022}. Investigations into human-AI interactive workflow for mathematical research also highlight the importance of human judgment in guiding AI~\cite{li2025advancing}. Vibe Reasoning introduces a distinct pattern where humans provide minimal, domain-agnostic guidance while AI performs substantive reasoning.

\textbf{Model Specialization and Orchestration.} Recent work has recognized that different models excel at different tasks~\cite{routerbench2024}. Our empirical findings on GPT-5 (exploration) versus Gemini 3 Pro (rigor) provide concrete evidence for this specialization and suggest that multi-model orchestration is essential for complex mathematical reasoning.

\section{Conclusion}

We have presented Vibe Reasoning, a human-AI collaborative paradigm that transforms AI's latent potential into manifested capability. Our key insight is that frontier AI models already possess the knowledge required to solve challenging problems---they simply do not know \textit{how}, \textit{what}, or \textit{when} to apply it. Through our case study on IMO 2025 P6, we demonstrated how lightweight human guidance can unlock this potential.

The four pillars of Vibe Reasoning each address specific AI limitations:

\textbf{AI as primary reasoner}: Capable foundation models provide the substantive mathematical knowledge and computational power, but require external guidance to deploy effectively.

\textbf{Socratic meta-prompts}: Generic, domain-agnostic human guidance supplies the missing judgment on how, what, and when---without revealing solutions or requiring domain expertise.

\textbf{Agentic grounding}: Python execution catches hallucinations and enables pattern discovery; file-based memory maintains coherence across sessions.

\textbf{Model orchestration}: Matching model strengths to task phases (GPT-5 for exploration, Gemini 3 Pro for proof) overcomes individual model limitations.

This paper represents work in progress. We are actively developing an automated Vibe Reasoning framework with three objectives: reducing human intervention, lowering expertise requirements, and improving reproducibility. By codifying effective meta-prompt patterns into reusable templates and decision rules, we aim to make Vibe Reasoning accessible to collaborators without high-level mathematical training, and to reducing human workload while maintaining solution quality. Also, replacing ad-hoc human judgment with systematic protocols enables reproducible problem-solving pipelines, facilitating scientific evaluation and iterative improvement of the methodology. We are also extending our evaluation to additional challenging benchmarks beyond IMO P6 to validate generality.

\section*{Acknowledgments}

We thank Sophia Zhang (zsophia@mit.edu) from MIT for providing valuable resources and suggestions during the early stages of this project. We also thank the AI systems (GPT-5, Gemini 3 Pro) for their computational efforts, and acknowledge that this work represents a collaborative achievement between human researchers and AI assistants.

\appendix

\section{Proof Trace Summary}
\label{appendix:proof-trace}

This appendix summarizes the complete reasoning trace, which contains over 100 AI turns across 2,941 lines. We highlight the key milestones and turning points in the solution process.

\subsection{Phase 1: Answer Discovery with GPT-5}

\textbf{Step 1: Problem Formalization.} GPT-5 first abstracted the grid covering problem into a permutation framework. A valid configuration corresponds to a permutation $\pi$ where row $i$ has its hole at column $\pi(i)$. The task becomes: find the minimum number of axis-aligned rectangles needed to cover all cells except the $n$ holes $(i, \pi(i))$.

\textbf{Step 2: Small Case Enumeration.} GPT-5 wrote a backtracking enumeration script to find exact minimum tile counts for small $n$. Initial results suggested $M(n) = 2(n-1)$, but this was quickly disproved when the human prompted verification---exact search showed $M(4) = 5 \neq 6$.

\textbf{Step 3: Pattern Discovery for Perfect Squares.} Following a human meta-prompt to focus on $n = k^2$ (since $2025 = 45^2$), GPT-5 analyzed optimal permutations for $n = 4, 9, 16, 25$ and discovered the ``residue block'' structure: positions are grouped by $i \mod k$ and blocks are arranged in descending order. This led to the conjecture:
$$M(k^2) = k^2 + 2k - 3$$
For $n = 2025 = 45^2$: $M(2025) = 2025 + 90 - 3 = 2112$.

\textbf{Step 4: Construction Verification.} GPT-5 verified the construction by generating explicit ASCII visualizations of optimal coverings for $n = 4, 9, 16$, confirming that the residue permutation achieves the conjectured minimum.

\subsection{Phase 2: Lower Bound Proof with Gemini 3 Pro}

\textbf{Step 5: Proof Strategy Selection.} Given the conjectured answer, Gemini 3 Pro proposed several proof strategies, and after several trials, it identified the Fooling Set method as the appropriate framework. A Fooling Set is a collection of cells where any axis-aligned rectangle containing two cells must also contain at least one hole.

\noindent$\ast$ \textbf{Note:} In the original full trace, the human seemed to directly instruct the AI to adopt the Fooling Set approach. This was to save time and computational cost after prior failure trials. AI can arrive at this framework through sequential exploration of self-generated strategy options without explicit guidance.

\textbf{Step 6: Fooling Set Construction.} Gemini 3 Pro developed the ``Adaptive Orthogonal Fanning'' strategy after self-verification, visualization and a limited number of hints:
\begin{enumerate}
    \item \textbf{Base layer} ($n$ cells): For each hole at $(i, \pi(i))$, select one adjacent cell in the ``outward'' direction.
    \item \textbf{LIS/LDS augmentation}: For positions in the Longest Increasing/Decreasing Subsequence, add cells in multiple directions.
\end{enumerate}
By the Erd\H{o}s-Szekeres theorem, $|LIS| + |LDS| \geq 2\sqrt{n}$, yielding the bound $n + 2\sqrt{n} - 3$ after boundary adjustments. Note that some hints provided during this phase were still problem-specific (e.g., suggesting LIS/LDS as coordinate axes); reducing such domain-specific guidance remains a key focus of our ongoing work.

\textbf{Step 7: Computational Verification.} The construction was verified with Python code on 100 random permutations for various $n$, achieving high success rate on random permutations and 100\% on the specific residue permutations.

\subsection{Fanning Strategy Visualization}
\label{appendix:fanning-visualization}

\begin{figure}[t]
\centering
\includegraphics[width=0.8\textwidth]{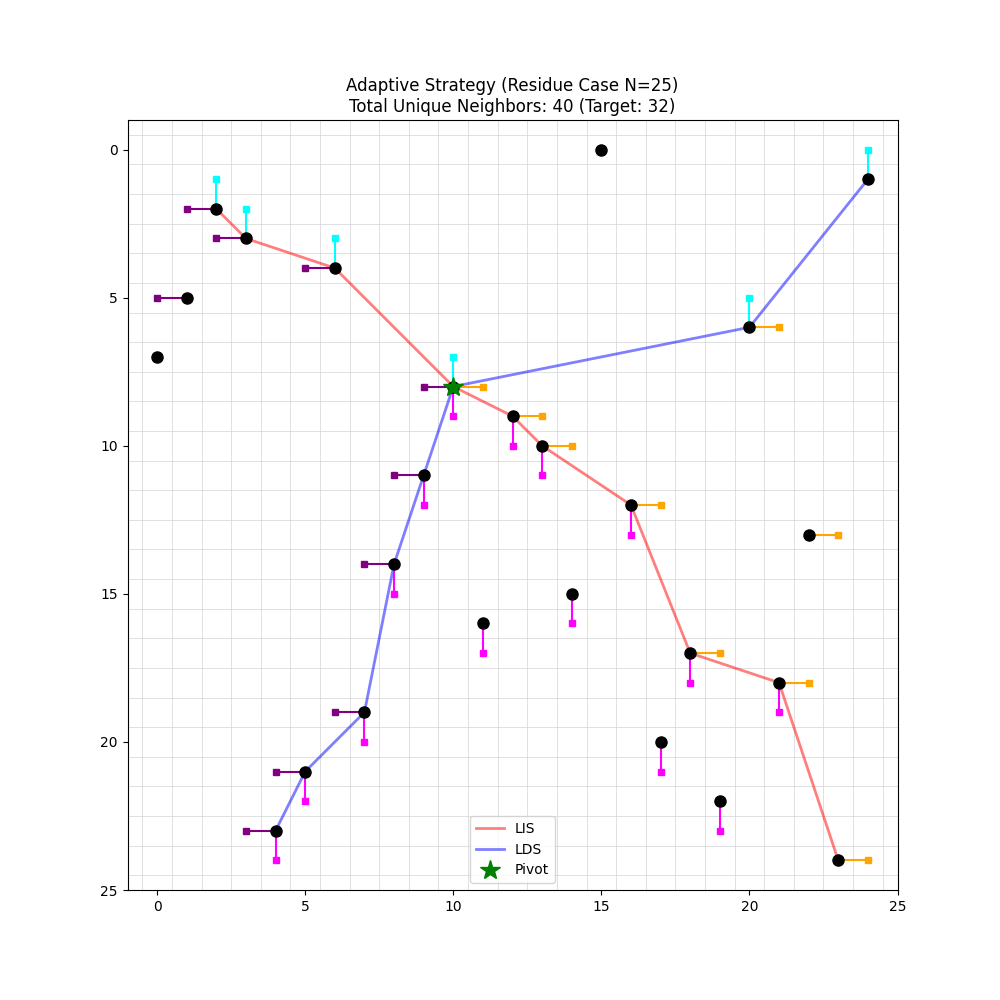}
\caption{Adaptive Orthogonal Fanning strategy for $n=25$. Black dots represent holes arranged in the residue permutation. The LIS (red line) and LDS (blue line) intersect at the pivot (green star). Marked cells fan outward. Total Fooling Set size: 40, exceeding the target bound of 32.}
\label{fig:fanning-visualization}
\end{figure}

Gemini 3 Pro generated a visualization of the Adaptive Orthogonal Fanning strategy for random permutation with $n=25$ ($k=5$), as shown in Figure~\ref{fig:fanning-visualization}. This shows the LIS (red) and LDS (blue) chains intersecting at the pivot point, with marked cells (fooling set elements) extending outward from each hole position. The elements which exceed the border of the grid are omitted.

\textbf{Step 8: Proof Formalization.} Gemini 3 Pro wrote the complete proof to a Markdown file (\texttt{formal\_proof\_lower\_bound.md}), providing detailed case analysis for the Fooling Set property.

\subsection{File System as Scratch Paper}

Throughout the reasoning process, the AI maintained several files that served as external memory, including but not limited to:
\begin{itemize}
    \item \texttt{summary.md}: Recorded enumeration results, pattern observations, and intermediate conjectures.
    \item \texttt{proof\_sketch.md}: Outlined proof strategies and key lemmas before formal writing.
    \item \texttt{conflict\_analysis.md}: Documented failed approaches constructing Fooling Sets.
    \item \texttt{construction\_visualization.md}: Described geometric intuitions and visual representations.
    \item \texttt{formal\_proof\_lower\_bound.md}: Final proof document with rigorous case analysis.
\end{itemize}
These files enabled coherent reasoning across the multi-hour interaction, allowing the AI to track what had been tried, what failed, and what remained to be done.

The full conversation trace and AI-generated Python, Markdown files are available at:\\ \url{https://github.com/Julius-Woo/Vibe-Reasoning/}.

\end{document}